\documentclass{article} 
\usepackage{iclr2025_conference,times}

\usepackage[utf8]{inputenc} 
\usepackage[T1]{fontenc}    

\usepackage{microtype}
\usepackage{graphicx}
\usepackage{booktabs}

\usepackage{amsmath,amsfonts,bm}

\def\eqref#1{equation~\ref{#1}}

\def\1{\bm{1}}

\DeclareMathAlphabet{\mathsfit}{\encodingdefault}{\sfdefault}{m}{sl}
\SetMathAlphabet{\mathsfit}{bold}{\encodingdefault}{\sfdefault}{bx}{n}

\usepackage{amsmath}
\usepackage{amssymb}
\usepackage{mathtools}
\usepackage{amsthm}
\usepackage{bm}

\usepackage{hyperref}
\usepackage[nameinlink]{cleveref}

\usepackage{multirow}           
\usepackage{amsfonts}           
\usepackage{nicefrac}
\usepackage{duckuments}         
\usepackage{thmtools,thm-restate}
\usepackage{enumitem}
\usepackage{xcolor,colortbl}
\usepackage{tikz}
\usepackage{tikz-cd}
\usepackage{caption}
\usepackage{subcaption}
\usepackage{listings}
\usepackage{gensymb}
\usepackage[inkscapelatex=false]{svg}
\usepackage{duckuments}
\usepackage{siunitx}
\usepackage{tikz}
\usepackage{tikz-cd}
\usepackage{xcolor}
\usepackage{colortbl}
\definecolor{lightblue}{rgb}{0.68, 0.85, 0.9}

\newcommand{\A}{\mathbb{A}}

\setlist[itemize]{leftmargin=2em} 
\setlist[enumerate]{leftmargin=2em} 

\usepackage{titletoc} 

\hypersetup{colorlinks=true,allcolors=[HTML]{6c5ce7}}

\title{Towards Mechanistic Interpretability of\\Graph Transformers via Attention Graphs}

\author{Batu El\thanks{\footnotesize Equal contribution. Work done in \href{https://www.cl.cam.ac.uk/teaching/2324/L65/}{Geometric Deep Learning} course at the University of Cambridge.}\\
Stanford University \\
\texttt{batuel@stanford.edu} \\
\And
Deepro Choudhury\footnotemark[1] \\
University of Oxford \\
\texttt{deepro.choudhury@stats.ox.ac.uk} \\
\AND
Pietro Liò \\
University of Cambridge \\
\texttt{pietro.lio@cl.cam.ac.uk}
\And
Chaitanya K. Joshi \\
University of Cambridge \\
\texttt{chaitanya.joshi@cl.cam.ac.uk}
}

\iclrfinalcopy 
\begin{document}

\maketitle

\begin{abstract}

We introduce \textit{Attention Graphs}, a new tool for mechanistic interpretability of Graph Neural Networks (GNNs) and Graph Transformers based on the mathematical equivalence between message passing in GNNs and the self-attention mechanism in Transformers. 
Attention Graphs aggregate attention matrices across Transformer layers and heads to describe how information flows among input nodes. 
Through experiments on homophilous and heterophilous node classification tasks, we analyze Attention Graphs from a network science perspective and find that: (1) When Graph Transformers are allowed to learn the optimal graph structure using all-to-all attention among input nodes, the Attention Graphs learned by the model do not tend to correlate with the input/original graph structure; and (2) For heterophilous graphs, different Graph Transformer variants can achieve similar performance while utilising distinct information flow patterns.
Open source code:
\href{https://github.com/batu-el/understanding-inductive-biases-of-gnns}{\small \texttt{github.com/batu-el/understanding-inductive-biases-of-gnns}}
\end{abstract}


\section{Introduction}

Graph Neural Networks (GNNs) and Graph Transformers (GTs) have emerged as core deep learning architectures behind several recent breakthroughs in physical and life sciences \citep{zhang2023artificial}, with AlphaFold \citep{jumper2021highly} being the most prominent example. 
Despite their remarkable success, these models remain largely `engineering artifacts', as noted by Demis Hassabis \citep{hassabis2024ai}, requiring dedicated tools to interpret their underlying mechanisms and advance our scientific understanding \citep{lawrence2024understanding}. 
While significant progress has been made in mechanistic interpretability of regular transformers in natural language processing \citep{elhage2021mathematical,bricken2023monosemanticity}, similar tools for GNNs and GTs for scientific applications are currently lacking.

\begin{figure}[t!]
    \centering
    \includegraphics[width=\linewidth]{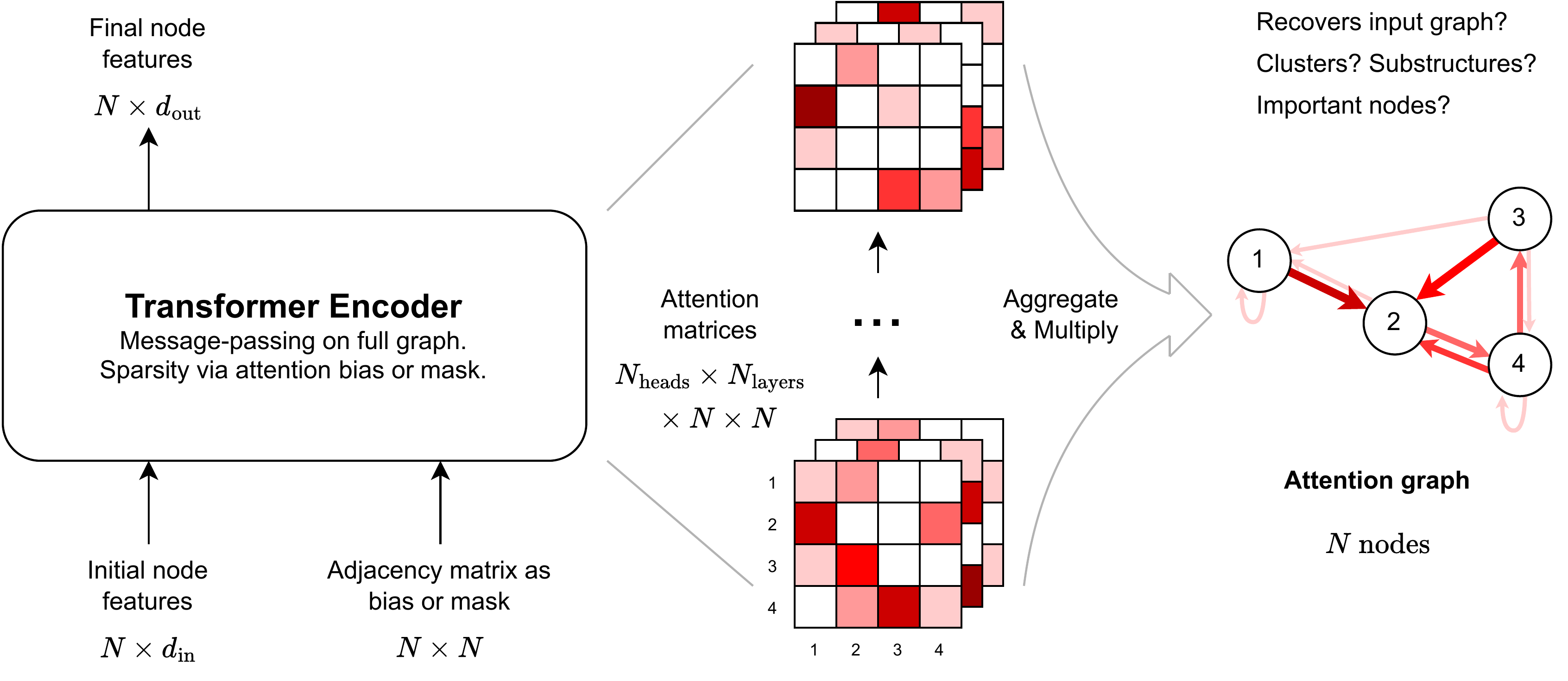}
    \caption{
    \textbf{Attention Graphs for mechanistic interpretability of GNNs and Graph Transformers. }
    \textit{Left:} Graph Neural Networks are equivalent to Transformers operating on fully connected graphs \citep{joshi2020transformers}.
    \textit{Middle:} The attention matrices at each layer and each head in the Transformer tell us how information flows among input tokens.  
    \textit{Right:} The attention matrices can be aggregated across layers and heads to construct a directed \textit{Attention Graph} of information flow in the GNN/Graph Transformer. 
    We can study Attention Graphs from a network science perspective to mechanistically understand the algorithms learned by GNNs and Graph Transformers.
    }
    \label{fig:pipeline}
\end{figure}

In this paper, we leverage the mathematical equivalence between message passing in GNNs and the self-attention mechanism in Transformers \citep{joshi2020transformers, vaswani2023attention} to investigate information flow patterns in these architectures. 
The core insight is that Graph Transformers produce two types of attention matrices: (1) matrices from different heads that capture distinct relationships between nodes, analogous to heterogeneous graphs; and (2) attention matrices across layers that represent how information flows through a dynamically evolving network over time. 
We introduce \textit{Attention Graphs}, a principled framework that aggregates these attention matrices into a unified representation of information flow among input nodes, as illustrated in \Cref{fig:pipeline}.
Through Attention Graphs, we utilize techniques from network science \citep{rathkopf2018network, krickel2023and} to analyze how learned information flow patterns in GTs relate to the underlying graph structure.
Based on the findings, we explore how different GT variants, which enforce graph structure to varying degrees, learn \textit{distinct} information flow patterns despite performing similarly.

Our experiments across multiple architectures and node classification tasks reveal two key findings. 
First, when architectures do not explicitly constrain attention to the underlying graph structure provided as input, the learned information flow patterns deviate significantly from the input graph topology - this holds for both homophilous and heterophilous graphs \citep{platonov2023a}. 
Secondly, on heterophilous graphs, different architectures can achieve similar performance while implementing \textit{distinct} algorithmic strategies, as evidenced by their unique patterns of information flow.
Moreover, we observe that certain nodes emerge as pivotal, disproportionately influencing predictions across the graph regardless of topological distance. 

Overall, we have introduced the first framework for mechanistic interpretability of GNNs and GTs based on network science.
While preliminary, this approach lays the foundation for future research toward understanding the algorithmic principles behind Transformers and attention-based models with the aim of unlocking deeper insights in their applications across the sciences.
Our code is available via:
\href{https://github.com/batu-el/understanding-inductive-biases-of-gnns}{\small \texttt{github.com/batu-el/understanding-inductive-biases-of-gnns}}.


\section{Preliminaries}

\textbf{Graph Neural Networks. }
A graph, $G = (V \ , \ E)$, is a mathematical structure that consists of a set of vertices $V$ representing entities and a set of edges $E \subseteq V \times V$ representing pairwise relationships between entities \citep{bronstein2021geometric}.
A Graph Neural Network (GNN) for node classification learns a function $f$ that maps nodes $V$ to an associated set of class labels $Y$ per node. 
Each node $i$ is represented by an input feature vector $x_i \in \mathbb{R}^{d}$, where $d$ is the feature dimension. 
We focus on graphs with binary relationships between nodes, characterized by an adjacency matrix $A \in \mathbb{R}^{n \times n}$, where $a_{ij} = 1$ if nodes $i$ and $j$ are connected, and $0$ otherwise. 
Thus, each graph is described by two matrices: the adjacency matrix $A$ and the node feature matrix $X \in \mathbb{R}^{n \times d}$, where $n$ is the number of nodes. 
The 1-hop neighbourhood of node $i$ is defined as $\mathcal{N}^1_i = \{j \mid a_{ij} = 1\}$. 
More generally, the $k$-hop neighborhood is $\mathcal{N}^k_i = \{j \mid (A^k)_{ij} \neq 0\}$, where $A^k$ denotes the $k$-th power of $A$.

GNNs are the standard toolkit for deep learning on graph-structured data, using graph topology to propagate and aggregate information between connected nodes \citep{velivckovic2023everything}.
Unlike feed-forward networks (standard multi-layer perceptrons) which process each node independently via $\text{MLP} (x_i) = y_i$, 
GNNs use both the node features $X$ and the graph structure $A$ simultaneously when predicting node classes. 
Formally, a GNN implements a permutation-equivariant function $f(X \ , \ A) = y$ such that $f(\textbf{P}X \ , \ \textbf{P}A\textbf{P}^{T}) = \textbf{P}y$ for any permutation matrix $\textbf{P}$. 
This permutation equivariance ensures the predictions for each node are invariant to node ordering.

\textbf{Message Passing. }
GNN layers are implemented via message-passing \citep{battaglia2018relational} where nodes iteratively update their representations by aggregating information from their neighbors. Formally, a node's representation $h_i^{\ell} \in \mathbb{R}^{d_{\text{model}}}$ at layer $\ell$ is updated to $h_i^{\ell +1}$ via:

\begin{align}
    h_i^{\ell+1} = \phi \Big( h_i^{\ell} \ , \ \bigoplus_{j \in \mathcal{N}^1_i} \psi \left( h_i^{\ell} \ , \ h_j^{\ell} \ , \ e_{ij} \right) \Big).
    \label{eq:mpnn}
\end{align}
Here, $\psi$ is a message function that determines how information flows between nodes $i$ and $j$ based on their representations and edge features $e_{ij}$, $\bigoplus$ is a permutation-invariant aggregation function (like sum or mean), and $\phi$ is a node-wise update function that combines the aggregated messages with each node's current representation.
$\psi$ and $\phi$ are typically implemented as multi-layer perceptrons (MLPs) with learnable weights shared across all nodes.
For the binary graphs in our experiments, edge features $e_{ij}$ simply equal the corresponding entries $a_{ij}$ in the adjacency matrix, though in general they can encode richer edge attributes.

Equation \ref{eq:mpnn} provides a general formulation that encompasses most widely-used GNN architectures, including Graph Convolutional Networks (GCNs) \citep{kipf2017semisupervisedclassificationgraphconvolutional}, Graph Isomorphism Networks (GINs) \citep{xu2019powerful}, and Message Passing Neural Networks (MPNNs) \citep{gilmer2017neural}.
We are particularly interested in attentional GNNs, where the message function $\psi$ is implemented via the self-attention operation \citep{Velickovic:2018we, brody2022attentive}.
In this case, the message function $\psi$ is a weighted sum of the representations of neighboring nodes, where the weights are computed using an attention mechanism.
\begin{align}
    \psi \left( h_i^{\ell}, h_j^{\ell}, e_{ij} \right) \ = \ \text{LocalAttention} \left( W^Q h_i^{\ell}, W^K h_j^{\ell}, W^V h_j^{\ell} \right)
    \ = \ \frac{\exp(Q_i^T K_j)}{\sum_{k \in \mathcal{N}^1_i} \exp(Q_i^T K_k)} \cdot V_j ,
    \label{eq:attention}
\end{align}
where $Q_i = W^Q h_i^{\ell} \ , \ K_j = W^K h_j^{\ell} \ , \ V_j = W^V h_j^{\ell}$ are the query, key, and value vectors, respectively, and $W^Q \ , \ W^K \ , \ W^V$ are learnable weight matrices.
In practice, we use multi-head attention to project the node representations into multiple subspaces and compute attention in parallel, followed by concatenation and a final linear transformation.

\textbf{Graph Transformers. }
GNNs and Transformers have deep mathematical connections \citep{joshi2020transformers}.
Transformers are attentional GNNs operating on fully-connected graphs, where self-attention models relationships between all pairs of input tokens \citep{vaswani2023attention}, i.e. graph nodes:
\begin{align}
    h_i^{\ell+1} \ &= \ \phi \Big( h_i^{\ell} \ , \ \bigoplus_{j \in V} \psi \left( h_i^{\ell} \ , \ h_j^{\ell} \ , \ e_{ij} \right) \Big) \ = \ \text{FFN} \Big( h_i^{\ell} + \sum_{j \in V} \psi \left( h_i^{\ell}, h_j^{\ell}, e_{ij} \right) \Big) ,
    \label{eq:mpnn_gt} \\
    \psi \left( h_i^{\ell}, h_j^{\ell}, e_{ij} \right) \ &= \ \text{GlobalAttention} \left( W^Q h_i^{\ell}, W^K h_j^{\ell}, W^V h_j^{\ell} \right)
    \ = \ \frac{\exp(Q_i^T K_j)}{\sum_{k \in V} \exp(Q_i^T K_k)} \cdot V_j .
    \label{eq:attention_gt}
\end{align}
This ability to attend to and gather information from all nodes allows Transformers to learn complex long-range dependencies without being constrained by a pre-defined graph structure or suffering from oversquashing bottlenecks \citep{di2023over}.

Conversely, GNNs are Transformers where self-attention is restricted to local neighborhoods \citep{buterez2024masked}, as formalized in \eqref{eq:attention}, which can be realized in \eqref{eq:mpnn_gt} by setting the message/attention weights to zero for any nodes $i$ and $j$ that are not connected in the graph, i.e. masking the attention.
These insights have given rise to Graph Transformers (GTs) that generalize Transformers for graph-structured data \citep{dwivedi2021generalization, muller2023attending}.
GTs aim to overcome oversquashing in GNNs by allowing global attention while still leveraging graph structure as an inductive bias \citep{rampavsek2022recipe}.

\textbf{Attention Matrices in Graph Transformers. }
Each GT layer performs a message passing operation as in \eqref{eq:mpnn_gt}, where the message function $\psi$ is implemented via the global self-attention operation in \eqref{eq:attention_gt}. 
We can thus define an attention matrix $\mathbb{A} \in \mathbb{R}^{n \times n}$ at each layer as a function of $H$ and $A$, where $H^{\ell} \in \mathbb{R}^{n \times d_{\text{model}}}$ contains the representation of a node from the graph in each of its rows and $A$ is the adjacency matrix.
A GT with $N_L$ layers and $N_H$ attention heads per layers will result in $N_L \times N_H$ attention matrices every time the model is run on an input.


\section{Attention Graphs Framework}

Processing graph-structured data through GNNs and GTs results in multiple attention matrices - one for each attention head in each layer. 
To understand these models, we need a principled approach to aggregate these matrices into a single \textit{Attention Graph} that captures the overall information flow among nodes. 
In this section, we first formalize a unified design space of GNNs and GTs based on their attention mechanisms, and then introduce our framework for constructing and analyzing Attention Graphs to reveal the algorithmic patterns learned by different architectures.

\subsection{Design space of Graph Transformers}
\label{sec:design-space}


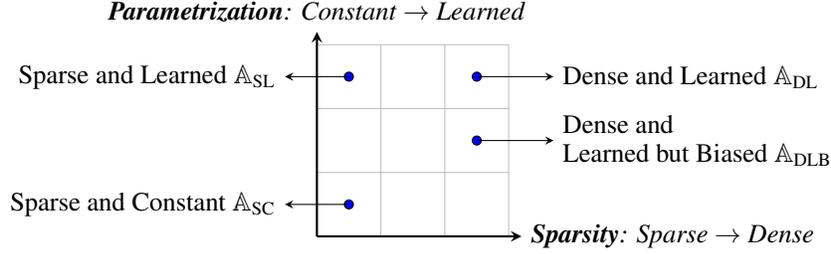
\begin{figure}[t!]
    \centering
    \begin{tikzpicture}[>=stealth,scale=0.85]
    
        \draw[step=1,lightgray,thin] (0,0) grid (3,3);
    
        \draw[->, thick] (0,0) -- (3.2,0)
            node[right]{\textit{\textbf{Sparsity}: Sparse} $\rightarrow$ \textit{Dense}};
        \draw[->, thick] (0,0) -- (0,3.2)
            node[above]{\textit{\textbf{Parametrization}: Constant} $\rightarrow$ \textit{Learned}};
    
        \coordinate (SC) at (0.5,0.5);  
        \coordinate (SL)  at (0.5,2.5);  
        \coordinate (DLB) at (2.5,1.5);  
        \coordinate (DL)  at (2.5,2.5);  
    
        \foreach \pt in {SC, SL, DLB, DL}{
            \draw[fill=blue] (\pt) circle (2pt);
        }
    
        \draw[->] (SC) -- (-0.5,0.5)
            node[left]{Sparse and Constant \(\mathbb{A}_{\text{SC}}\)};
    
        \draw[->] (SL) -- (-0.5,2.5)
            node[left]{Sparse and Learned \(\mathbb{A}_{\text{SL}}\)};
    
        \draw[->] (DLB) -- (3.7,1.5)
            node[right, align=left]{Dense and \\ Learned but Biased \(\mathbb{A}_{\text{DLB}}\)};
    
        \draw[->] (DL) -- (3.7,2.5)
            node[right]{Dense and Learned \(\mathbb{A}_{\text{DL}}\)};
    
    \end{tikzpicture}
    \caption{\textbf{Design space of Graph Transformers} based on two key dimensions: (1) sparsity of attention (sparse vs. dense) and (2) parametrization of attention (constant vs. learned).}
    \label{fig:attention-types}
\end{figure}
    

Having established the connection between GTs and GNNs through the attention mechanism and resulting attention matrices, we can formalise a spectrum of  architectures based on two key dimensions: the \textit{parametrization} and \textit{sparsity} of the attention matrix $\mathbb{A}$.
The parametrization dimension refers to whether the attention matrix is fixed or learned, while the sparsity dimension refers to whether the attention matrix is restricted to the neighborhood of the node or is allowed to be dense/global.
\Cref{fig:attention-types} visualizes these dimensions.

\textbf{Sparse and Constant (SC):} The attention matrix is sparse and fixed, with attention coefficients inversely proportional to the square root of the degree of the node being attended to, recovering GCNs \citep{kipf2017semisupervisedclassificationgraphconvolutional}. The layer-wise update equation is:
    \begin{align}
        h_i^{\ell+1} \ = \ \text{FFN} \Big( h_i^{\ell} \ + \ \sum_{j \in \mathcal{N}_i^1} \frac{1}{\sqrt{d_i d_j}} \cdot V_j \Big) .
    \end{align}

\textbf{Sparse and Learned (SL):} The attention matrix is sparse but learned, with attention coefficients learned as a function of the node features, recovering GATs \citep{Velickovic:2018we, brody2022attentive}.
    The layer-wise update equation is:
    \begin{align}
        h_i^{\ell+1} \ = \ \text{FFN} \Big( h_i^{\ell} \ + \ \sum_{j \in \mathcal{N}_i^1} \frac{\exp(Q_i^T K_j)}{\sum_{k \in \mathcal{N}^1_i} \exp(Q_i^T K_k)} \cdot V_j \Big) .
    \end{align}

\textbf{Dense and Learned but Biased (DLB):} The attention matrix is dense and learned, but biased to be inversely proportional to the shortest path length between the nodes, recovering the Graphormer \citep{ying2021do}.
    The layer-wise update equation is:
    \begin{align}
        h_i^{\ell+1} \ = \ \text{FFN} \Big( h_i^{\ell} \ + \ \sum_{j \in V} \frac{\exp(Q_i^T K_j \ + \ b_{ij})}{\sum_{k \in V} \exp(Q_i^T K_k  \ + \ b_{ij})} \cdot V_j \Big) , 
    \end{align}
    where the bias term $b_{ij} = 1 / \text{shortest path length}$ between nodes $i$ and $j$ encourages the model to attend to nodes that are closer in the graph without enforcing a strict neighborhood structure.

\textbf{Dense and Learned (DL):} The attention matrix is dense and learned in an unbiased manner, akin to the Transformer attention mechanism \citep{vaswani2023attention}.
    This is the most flexible model that allows each node to attend to every other node in the graph without any graph structure bias.
    The layer-wise update equation is:
    \begin{align}
        h_i^{\ell+1} \ = \ \text{FFN} \Big( h_i^{\ell} \ + \ \sum_{j \in V} \frac{\exp(Q_i^T K_j)}{\sum_{k \in V} \exp(Q_i^T K_k)} \cdot V_j \Big) , 
    \end{align}
All these models can be implemented using the same Transformer architecture, with the only difference being the attention mechanism used in the message passing operation.
In practice, we use the PyTorch module {\small \texttt{torch.nn.TransformerEncoder}} to implement all models, with the main difference being the attention mask and/or bias used \citep{buterez2024masked, dong2024flex}. 


\subsection{Aggregating Attention Across Heads and Layers}
\label{sec:agg-attention}
\label{sec:attention-across-heads}
\label{sec:attention-across-layers}

To understand how information flows within GNNs and Graph Transformers, we need to combine multiple attention matrices: (1) attention matrices from different heads, which can be viewed as capturing different types of relationships between nodes, similar to heterogeneous graphs; and (2) attention matrices across layers, which represent how information flows through a dynamically evolving network over time. This section presents our framework for aggregating these matrices into a single Attention Graph that reveals the overall information flow patterns in the model.

\textbf{Aggregating attention across heads. }
Attention matrices across heads can be viewed as representing different types of relationships between nodes, analogous to edge types in heterogeneous graphs. 
To enable systematic analysis, we need to combine these relationships into a single homogeneous graph whose edges capture the aggregate information flow from all attention heads at a given layer.
We first examine whether different heads learn similar or distinct attention patterns in Appendix \Cref{fig:Heads1L2H}, which shows pairwise correlations between attention values learned by different heads in 1-layer 2-head variants of our models defined in \Cref{sec:design-space}.
For homophilous datasets, SL models exhibit strong positive correlations between heads, indicating they converge on similar attention patterns.
This correlation remains positive but weaker for heterophilous datasets in both SL (row 1) and DLB models (row 2).
DL models (row 3) show the weakest correlations, though still consistently positive.
Crucially, we never observe negative correlations between heads, suggesting they learn complementary rather than competing patterns.

These intuitive observations motivate aggregatingacross heads through simple averaging: $\A_{\text{Agg.}} = \frac{1}{N_H} \sum_{i=1}^{N_H} \A_{Hi},$ where $N_H$ is the number of heads and $\A_{Hi}$ is the attention matrix for head $i$.
While straightforward, this averaging approach effectively captures the overall information flow patterns learned by multi-head attention without losing important signals from any individual head.


\begin{figure}[t!]
    \centering
    \includegraphics[width=0.9\linewidth]{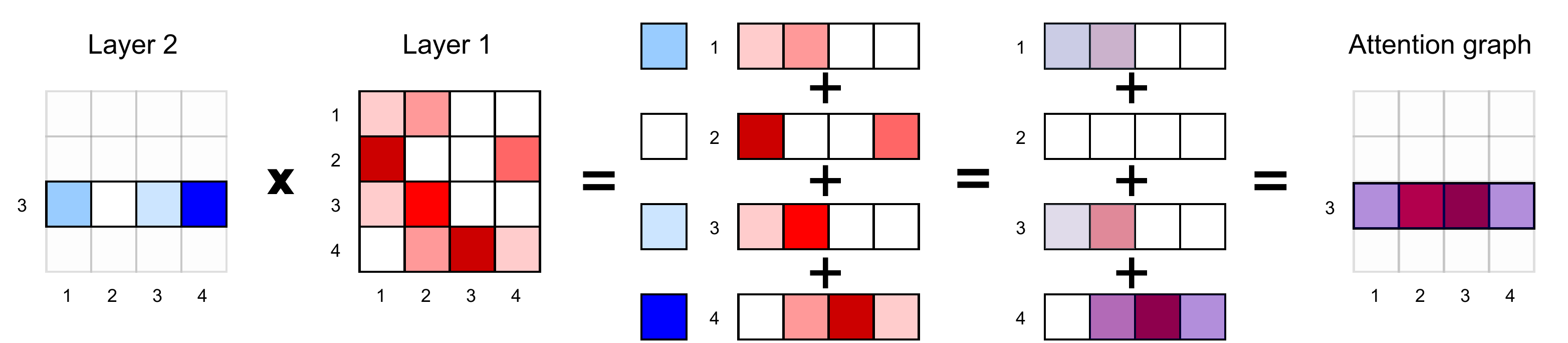}
    \caption{
    \textbf{Aggregating attention across layers by matrix multiplication. } 
    Attention matrices from successive layers are combined to capture indirect information flow. For node $i$, row $i$ in the attention matrix $\mathbb{A}_{L_2}$ represents how much it attends to each intermediate node $j$. Each row $j$ in $\mathbb{A}_{L_1}$ captures how those intermediate nodes attend to other nodes $k$. Matrix multiplication $\mathbb{A}_{L_2}\mathbb{A}_{L_1}$ combines these patterns, revealing how node $i$ indirectly attends to node $k$ through intermediate nodes $j$.
    }
    \label{fig:matmul}
\end{figure}


\textbf{Aggregating attention across layers. }
Attention matrices across layers capture how information flows through the network over multiple message passing steps, forming dynamic graphs that evolve temporally.
Our goal is to aggregate these matrices into a single Attention Graph that captures the overall flow of information across all layers of the model.
We analyzed the correlation between attention patterns learned in different layers in Appendix \Cref{fig:layers2L1H}, which shows pairwise correlations between attention values learned in first and second layers of 2-layer 1-head models.
We observed positive but weaker correlations compared to across-head patterns. This suggests that simple averaging may not be sufficient for combining attention across layers, necessitating a more nuanced approach that captures how information propagates through the network.

We propose using matrix multiplication of attention matrices from successive layers to model the information flow from successive layers (illustrated in \Cref{fig:matmul}):
\begin{align}
    \A_{\mathrm{Agg.}} = \A_{L_2} \A_{L_1},
\end{align}
This formulation elegantly captures indirect attention patterns: if node $i$ attends to node $j$ in layer 2, it indirectly attends to all nodes that $j$ attended to in layer 1. Mathematically, row $i$ of $\A_{\mathrm{Agg.}}$ represents a linear combination of rows in $\A_{L_1}$, weighted by attention coefficients in row $i$ of $\A_{L_2}$. This multiplication operation naturally models multi-hop information flow - for example, if node $j$ attends to $k$ in layer 1 and node $i$ attends to $j$ in layer 2, the matrix product captures the indirect flow of information from $k$ to $i$ through the intermediate node $j$.

\textbf{Constructing the Attention Graph. }
After performing a forward pass through the model, we obtain attention matrices $\A_{\ell h}$ for each layer $\ell$ and head $h$. To construct the Attention Graph, we first aggregate attention across heads at each layer to obtain layer-wise attention matrices $\A_{\ell}$, then multiply these matrices sequentially across layers to construct the final aggregate attention matrix $\A_{\mathrm{Agg.}}$.
This process captures the overall information flow patterns learned by the model, revealing how information propagates through the network over time.


\section{Experiments}
\label{sec:experiments}

\subsection{Experimental Setup}

\textbf{Datasets. }
We evaluate our framework on 7 node classification datasets with varying levels of homophily, with further details in \Cref{apdx:data}.
The datasets include:
(1) \textit{Citation Networks (Homophilous):} Cora and Citeseer \citep{Yang2016revisiting}, where nodes represent scientific papers (with bag-of-words features from abstracts), edges represent citations, and classes are research topics.
(2) \textit{Wikipedia Networks (Heterophilous):} Chameleon and Squirrel \citep{Rozemberczki2019}, where nodes are Wikipedia articles (with noun presence features), edges are hyperlinks, and classes are based on monthly traffic.
(3) \textit{University Webpages (Heterophilous):} Cornell, Texas, and Wisconsin from WebKB \citep{Pei2020Geom-GCN:}, where nodes are university webpages (with bag-of-words features), edges are hyperlinks, and classes are webpage categories.


\textbf{Models. }
We experiment with four variants of Graph Transformers defined in \Cref{sec:design-space}: SC, SL, DLB, and DL.
These models span a spectrum of attention mechanisms, from sparse and fixed to dense and learned, allowing us to systematically analyze the impact of different inductive biases on the information flow patterns learned by the model.
We use the Transformer Encoder module in PyTorch to implement all models, with the only difference being the attention mask and/or bias used.
We experiment with number of layers $N_L \in \{1, 2\}$ and number of heads $N_H \in \{1, 2\}$, resulting in 4 model variants for each dataset (described subsequently).
We set the hidden dimension $d_{\text{model}} = 128$ for all models, resulting in node representations $H^{\ell} \in \mathbb{R}^{n \times 128}$.

\textbf{Node classification performance}
Before analyzing the mechanistic interpretability of our models, we first evaluate their performance on node classification tasks.
While achieving state-of-the-art performance is not our primary goal, we verify that our models are competitive with existing approaches.
As shown in Appendix \Cref{tab:acc-model_comparison}, we observe distinct performance patterns across different graph types:
(1) On homophilous graphs (Cora, Citeseer), models that restrict attention to local neighborhoods (SC, SL) achieve the highest accuracy.
(2) On heterophilous graphs (Cornell, Texas, Wisconsin), removing neighborhood constraints (DL) leads to better performance, suggesting the importance of long-range interactions.
(3) On moderately homophilous graphs (Chameleon, Squirrel), biasing attention towards neighbors without strict restrictions (DLB) achieves optimal results.
Notably, these performance patterns remain consistent across different model configurations, with minimal variation when increasing the number of attention heads or layers.


\begin{figure}[t!]
    \centering
    \includegraphics[width=\linewidth]{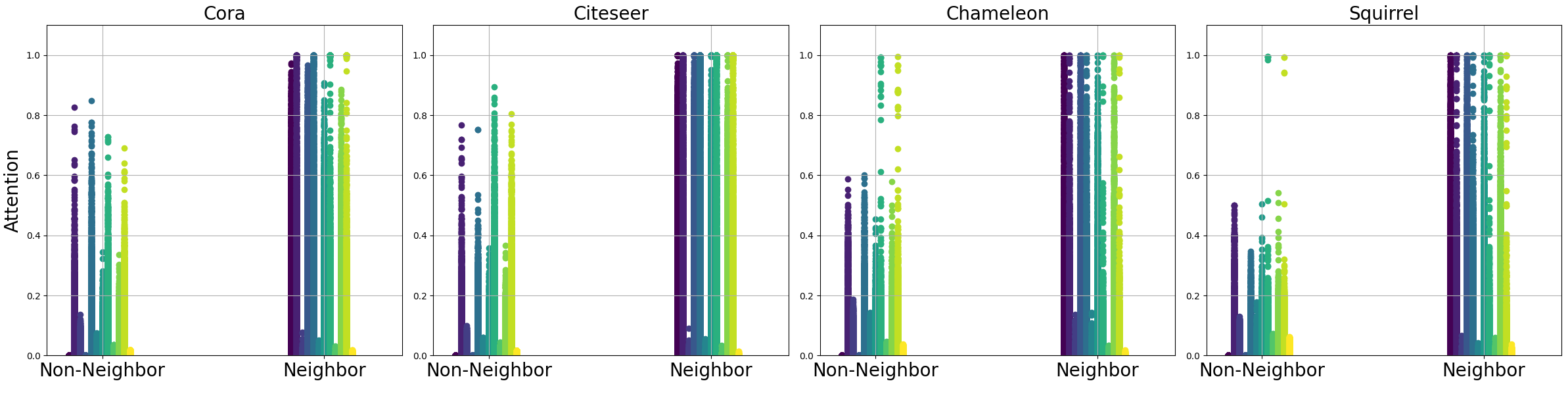}
    \includegraphics[width=\linewidth]{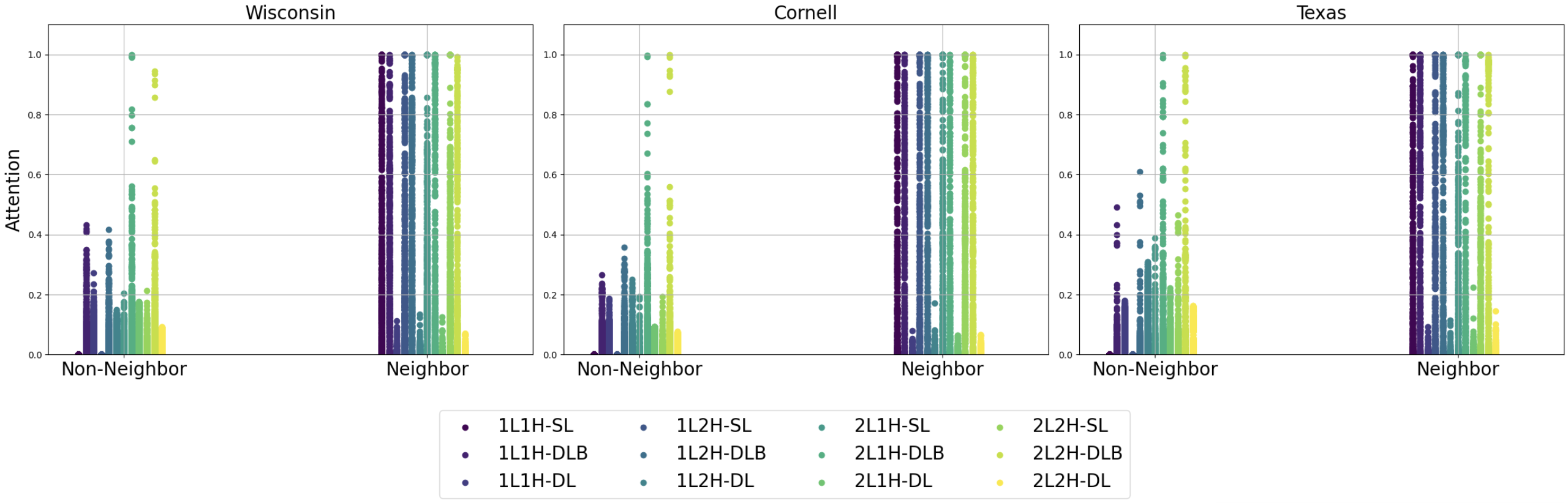}
     \caption{\textbf{Distribution of attention between neighbors and non-neighbors across different Graph Transformer architectures.} For SL (Sparse Learned), DLB (Dense Learned with Bias), and DL (Dense Learned) models, we visualize the attention patterns for four configurations: 1-layer 1-head, 1-layer 2-head, 2-layer 1-head, and 2-layer 2-head. Each point represents the weight of attention paid by a node in the aggregated Attention Graph, and whether it attends to a neighbor or non-neighbor from the input graph.
     DLB models mostly attend to neighbors, while DL models distribute attention more uniformly between neighbors and non-neighbors.
     }
    \label{fig:attention-to-neighbor}
\end{figure}


\subsection{How do Graph Transformers distribute attention?}

\Cref{fig:attention-to-neighbor} visualizes how different GT variants distribute attention between neighboring and non-neighboring nodes. In single-layer SL models, nodes can only attend to their immediate neighborhood due to the architectural constraint. However, in two-layer models, attention to non-neighbors emerges naturally through information propagation across the network - a phenomenon effectively captured by our attention aggregation framework. DLB models, with their graph-biased attention, concentrate most of their attention within local neighborhoods. In contrast, DL models distribute attention roughly uniformly between neighbors and non-neighbors, as quantified in Appendix \Cref{tab:attention-ratio} through the ratio of average attention paid to non-neighbors versus neighbors. To examine these patterns more granularly, Appendix \Cref{fig:attention-nhop-distribution} breaks down attention distribution by n-hop distances for 1-layer 1-head models. This analysis reveals that DLB models exhibit strong locality bias with attention decaying sharply with hop distance, while DL models maintain relatively uniform attention across all $k$-hop neighborhoods, suggesting fundamentally different information aggregation strategies.


\subsection{Do Graph Transformers recover input graph structure?}

Next we check whether the learned attention in our dense models matches the underlying graph structure. 
We construct quasi-adjacency matrices by thresholding Attention Graphs to recover binary connectivity patterns, as described in \Cref{apdx:recovering-graph-structure}.
\Cref{tab:F1} compares the quasi-adjacency matrices against original adjacency matrices using F1-scores. Low F1-scores (<4\%) for DL models across all datasets indicate their attention patterns do not reflect the input graph structure. In contrast, DLB models show moderate to high F1-scores (28-86\%), suggesting they partially recover graph connectivity through biased attention. Multi-head models generally achieve higher F1-scores, likely because averaging across heads reduces noise in learned attention patterns. However, for both architectures, F1-scores decrease with model depth, indicating that deeper layers develop more complex information flow patterns beyond the original graph structure.


\label{sec:recovering-graph-structure}
\begin{table}[t!]
     \caption{\textbf{Comparing graph structure recovery between DLB and DL models} using F1-scores between original adjacency matrices and thresholded attention matrices. 
    Higher F1-scores indicate better preservation of the original graph structure in the model's learned attention patterns. 
    DL models consistently achieve lower F1-scores compared to DLB models, suggesting they do not recover the input graph structure.
    }
    \label{tab:F1}
    \centering
    \begin{tabular}{lcccccccc}
    \toprule
    & \multicolumn{2}{c}{1L1H} & \multicolumn{2}{c}{1L2H} & \multicolumn{2}{c}{2L1H} & \multicolumn{2}{c}{2L2H} \\
    & DLB & DL & DLB & DL & DLB & DL & DLB & DL \\
    \midrule
    Cora       & 46.75 & 0.13 & 61.58 & 0.38 & 36.58 & 0.21 & 37.94 & 0.32 \\
    Citeseer   & 28.24 & 0.22 & 41.51 & 0.69 & 33.70 & 0.23 & 39.02 & 0.13 \\
    Chameleon  & 51.01 & 0.76 & 58.63 & 0.72 & 28.32 & 1.24 & 33.98 & 0.39 \\
    Squirrel   & 84.11 & 0.07 & 86.31 & 0.07 & 49.22 & 0.43 & 50.10 & 0.51 \\
    Cornell    & 57.44 & 0.21 & 61.38 & 0.85 & 43.51 & 1.25 & 48.95 & 1.04 \\
    Texas      & 58.37 & 0.61 & 58.31 & 1.22 & 44.83 & 3.69 & 47.14 & 2.95 \\
    Wisconsin  & 58.06 & 1.48 & 58.78 & 0.95 & 41.94 & 1.06 & 41.12 & 1.47 \\
    \bottomrule
    \end{tabular}
\end{table}



\subsection{A Closer Look at Quasi-Adjacency Matrices}
\label{sec:a-closer-look}

Finally, we analyze the quasi-adjacency matrices to understand information flow patterns across different architectures. \Cref{fig:HM} reveals three key findings:

\textbf{1. Strong Self-Attention in Dense and Learned but Biased Models:} The quasi-adjacency matrices learned by DLB models show prominent diagonal patterns, particularly in two-layer configurations, indicating nodes primarily attend to themselves. This suggests these models may solve the classification task by focusing on initial node features rather than extensively aggregating information from neighboring nodes, especially for heterophilous graphs where local structure is less informative.

\textbf{2. Reference Nodes in Dense and Learned Models:} In contrast, DL models consistently exhibit distinct vertical patterns in their quasi-adjacency matrices across all datasets, suggesting the emergence of "reference nodes" that receive high attention from all other nodes. These patterns become more pronounced in two-layer models and larger heterophilous graphs like Chameleon and Squirrel. We hypothesize that DL models implement a classification algorithm based on comparing nodes against these learned reference nodes rather than relying on local graph structure.
This finding is also supported by the benefits of including global/virtual nodes or register tokens in GTs \citep{gilmer2017neural, darcet2024vision}.

\textbf{3. Distinct Algorithms, Similar Performance:} Most notably, while DLB and DL models achieve comparable accuracy on heterophilous tasks (\Cref{tab:acc-model_comparison}), their attention patterns reveal fundamentally different algorithmic strategies. This finding challenges the common practice of evaluating models solely based on accuracy metrics and highlights the importance of analyzing internal model behavior to understand learned computational strategies.


\begin{figure}[t!]
    \centering
    \includegraphics[width=\linewidth]{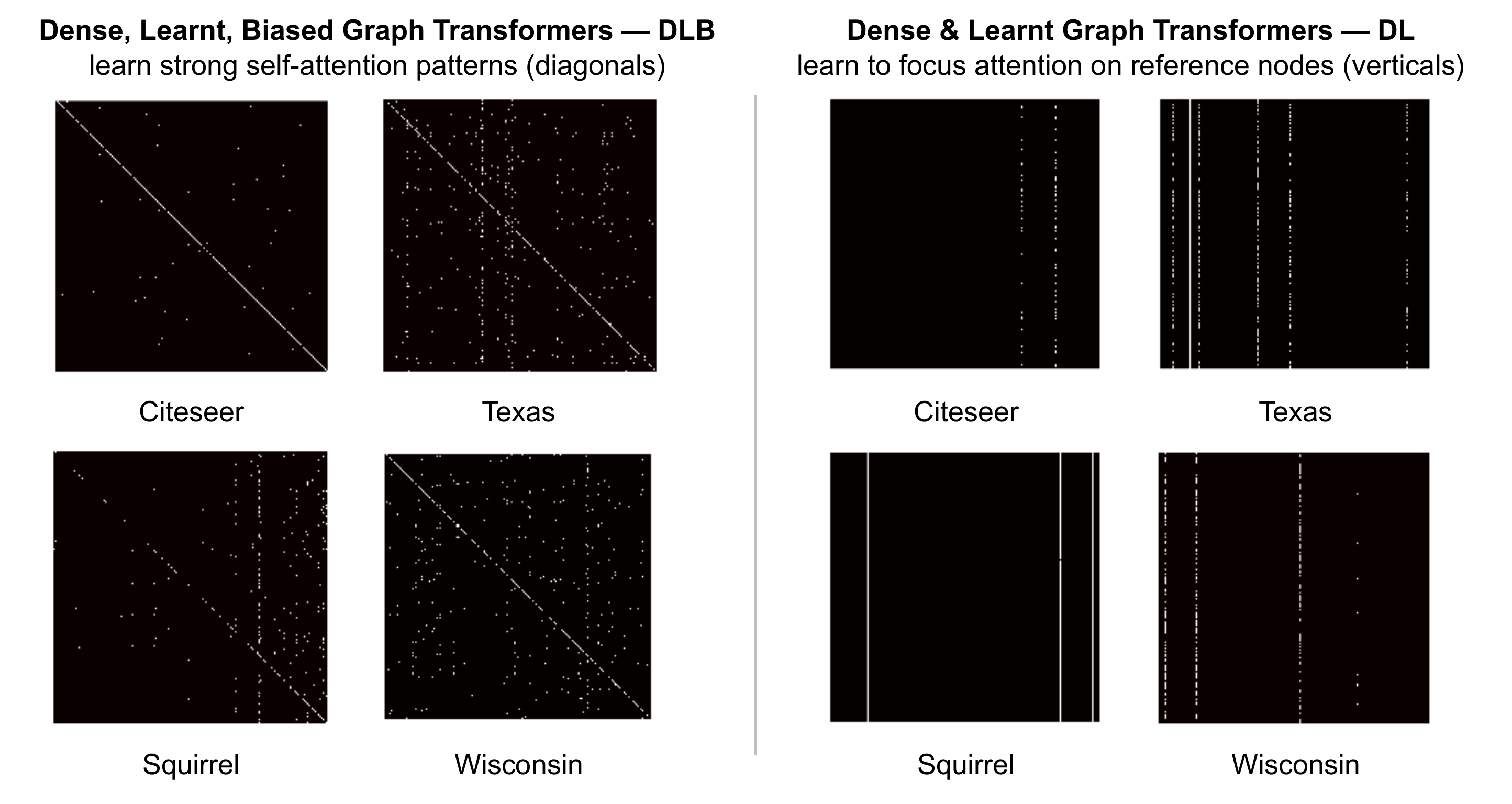}
     \caption{
        \textbf{Different graph inductive biases lead to distinct algorithmic strategies.}
        We plot quasi-adjacency matrices derived from Attention Graphs  for DLB and DL models across different datasets for the 2-layer 2-head configuration.
        Black squares indicate no edges in the thresholded Attention Graph, while white squares indicate edges.
        DLB models exhibit strong self-attention patterns (diagonal lines), suggesting they focus on initial node features rather than aggregating information from neighbors.
        DL models develop reference nodes (vertical lines) that receive high attention from all other nodes, suggesting a classification algorithm based on comparing nodes against these references.
        See \Cref{fig:HM} for all model configurations and datasets.
     }
    \label{fig:mechanisms}
\end{figure}


\section{Related Work}

Our work bridges GNN explainability and mechanistic interpretability of Transformers, aiming to understand how information flows through these architectures during inference from the perspective of graph theory and network science \citep{rathkopf2018network,krickel2023and}.

\textbf{GNN Explainability.} Early work on explaining GNNs focused on identifying influential subgraphs for specific predictions \citep{ying2019gnnexplainer}. This spawned several approaches including concept-based methods \citep{magister2021gcexplainer}, counterfactual explanations \citep{lucic2022cfgnnexplainer}, and generative explanations \citep{Yuan_2020}. While valuable, these methods primarily analyze input-output relationships rather than internal model dynamics. Other work has investigated physical laws learned by GNNs using symbolic regression \citep{cranmer2020discovering}, but a systematic framework for understanding information flow in GNNs remains lacking.

\textbf{Mechanistic Interpretability of Transformers.} Recent advances in mechanistic interpretability aim to reverse-engineer neural networks into human-understandable components \citep{olah2022mechanistic,elhage2021mathematical}. This has led to breakthroughs in understanding model features \citep{olah2017feature,elhage2022toy}, identifying computational circuits \citep{nanda2023progress,cammarata2020thread:}, and explaining emergent behaviors \citep{barak2023hidden,wei2022emergent}. These insights have practical benefits - enabling better out-of-distribution generalization \citep{mu2021compositional}, error correction \citep{hernandez2022natural}, and prediction of model behavior \citep{meng2022locating}. 
Our work aims to extend the principles of mechanistic interpretability to GNNs by leveraging their mathematical connection to Transformers. 
A complementary effort is emerging around mechanistic interpretability of biological language models \citep{Zhang2024, simon2024interplm, adams2025from}, sharing our goal of extracting scientific insights from AI systems trained on structured scientific data. 


\section{Discussion}

In this paper, we developed a framework to mechanistically interpret GNNs and Graph Transformers by analyzing their attention patterns from the perspective of network science.
While previous work on Transformer circuits focused on discrete feature interactions \citep{bricken2023monosemanticity,meng2022locating}, our approach captures continuous information flow patterns between nodes through Attention Graphs. Through this lens, we demonstrated that architectures with different graph inductive biases can achieve similar performance while implementing distinct algorithmic strategies. This observation challenges the common practice of evaluating models solely based on accuracy metrics and highlights the need for more holistic evaluation approaches.

Our framework has important limitations to address in future work. First, while matrix multiplication effectively models indirect attention flow, it may not capture non-linear interactions from activation functions between layers. Second, aggregating heterogeneous attention patterns across heads and temporal patterns across layers into a single matrix may oversimplify complex model dynamics. Additionally, our preliminary experiments are currently limited to small models (up to 2 layers, 2 heads) and node classification tasks where homophily analysis provides clear insights into the importance of graph structure.
Future work should extend the Attention Graph framework to more complex graph architectures and inductive tasks to develop a more complete understanding of information flow in Graph Transformers. 
Attention Graphs opens up numerous directions for applying network science and graph theory to analyze Transformers, including spectral analysis, community detection, and information flow dynamics. These tools could reveal deeper insights into the emergent computational strategies learned by these models.



\section*{Acknowledgements}

We would like to thank Petar Veličković, Dobrik Georgiev, and Jonas Jürß for helpful discussions. BE is supported by the Knight-Hennessy Scholarship. DC is supported by the Oxford-Radcliffe PhD Scholarship and an EPSRC PhD Studentship for the StatML Centre for Doctoral Training.
CKJ is supported by the A*STAR Singapore National Science Scholarship (PhD) and the Qualcomm Innovation Fellowship.


{
\small
\bibliography{iclr2025_conference}
\bibliographystyle{iclr2025_conference}
}

\appendix

\newpage

\section{Datasets and Homophily Metrics}
\label{apdx:data}

We evaluate our framework on seven node classification datasets with varying levels of homophily. \Cref{tab:dataset-summary} summarizes the datasets and homophily metrics used in this paper.
Homophily measures the likelihood of nodes with the same class label to connect with each other, providing insights into the underlying graph structure and the relationships between nodes.


\begin{table}[h!]
    \caption{\textbf{Node classification datasets studied in this paper.} Homophily metrics range from -100 to 100, with higher values indicating stronger homophily (nodes of same class are more likely to connect). Number of classes indicates the number of unique labels per node.}
    \label{tab:dataset-summary}
    \centering
    \small
    \begin{tabular}{lrrrrrrr}
    \toprule
    \textbf{Metric} & \textbf{Cora} & \textbf{Citeseer} & \textbf{Chameleon} & \textbf{Squirrel} & \textbf{Cornell} & \textbf{Texas} & \textbf{Wisconsin} \\
    \midrule
    \text{Node Homophily}       & 82.5  & 70.6  & 10.4   & 8.9    & 10.6   & 6.5   & 17.2 \\
    \text{Edge Homophily}       & 81.0  & 73.6  & 23.5   & 22.4   & 13.1   & 10.8  & 19.6 \\
    \text{Adjusted Homophily}   & 77.1  & 67.1  & 3.3    & 0.7    & -21.1  & -25.9 & -15.2 \\
    \text{Number of Nodes}      & 2708  & 3327  & 2277   & 5201   & 183    & 183   & 251 \\
    \text{Number of Edges}      & 10556 & 9104  & 36101  & 217073 & 298    & 325   & 515 \\
    \text{Number of Classes} & 7     & 6     & 5      & 5      & 5      & 5     & 5 \\
    \bottomrule
    \end{tabular}
    \end{table}
    

\textbf{Node Homophily} computes, for each node, the fraction of its neighbors that share its class label, then averages across all nodes \citep{Pei2020Geom-GCN:}:
\[
h_{\text{node}} = \frac{1}{|V|} \sum_{v \in V} 
    \frac{ \left| \{ u \in N(v) : y_u = y_v \} \right| }{ d(v) },
\]
where $V$ is the node set, $N(v)$ contains the neighbors of node $v$, $y_v$ denotes $v$'s class label, and $d(v)$ is $v$'s degree.

\textbf{Edge Homophily} calculates the fraction of edges that connect nodes of the same class \citep{zhu2020homophily, elhaijamixhop}:
\[
h_{\text{edge}} = \frac{ \left| \{ \{u,v\} \in E : y_u = y_v \} \right| }{|E|},
\]
where $E$ is the edge set. While intuitive, this metric can be misleading when class distributions are imbalanced.

\textbf{Adjusted Homophily} addresses class imbalance by comparing observed homophily against expected homophily given class proportions \citep{newman2003mixing, platonov2023characterizing}:
\[
h_{\text{adj}} 
  = h_{\text{edge}} 
     - \frac{\sum_{k=1}^{K} \bar{p}(k)^2}{1 - \sum_{k=1}^{K} \bar{p}(k)^2},
\]
where $\bar{p}(k)$ represents the fraction of nodes in class $k$, and $K$ is the total number of classes. This formula normalizes the difference between observed edge homophily $h_{\text{edge}}$ and expected homophily $\sum_{k=1}^K \bar{p}(k)^2$, providing a more balanced measure of homophily in graphs with uneven class distributions.


\section{Quasi-Adjacency Matrix}
\label{apdx:recovering-graph-structure}

To investigate whether learned attention patterns reflect the original graph structure, we convert aggregate Attention Graphs into binary "quasi-adjacency" matrices through thresholding. Following \citet{knyazev2019understanding}, we set entries above a threshold to 1 and below to 0. The threshold is chosen such that the number of edges in the quasi-adjacency matrix approximately matches the original graph, helping balance precision and recall in our structure recovery analysis. We determine this threshold through grid search over [0,1] with 0.001 increments. This approach allows fair comparison between learned attention patterns and input graph structure while avoiding bias from overly sparse or dense quasi-adjacency matrices.

\newpage 

\section{Intuition for attention aggregation across heads and layers}

\begin{figure}[h!]
    \centering
    \includegraphics[width=\textwidth]{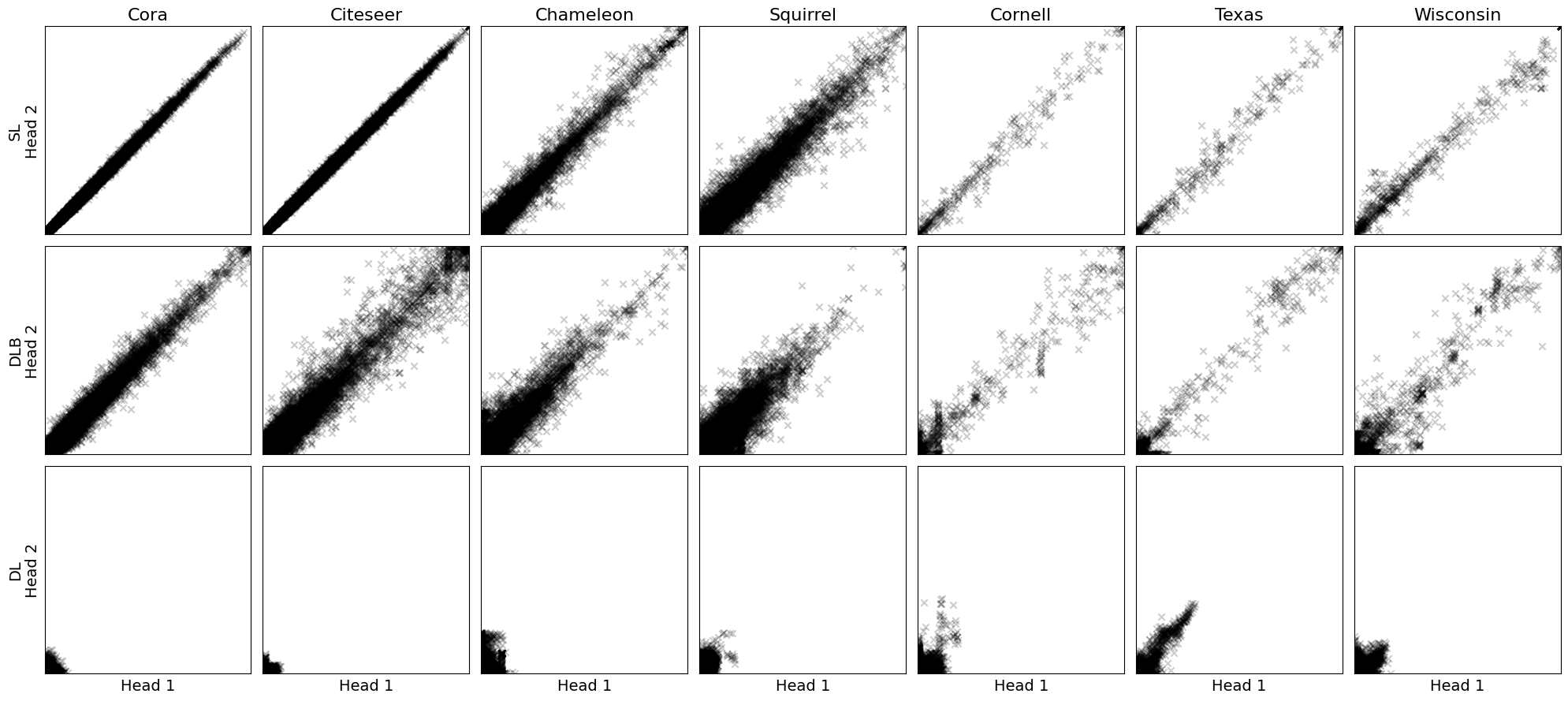}
    \caption{Comparison of attention patterns between heads in 1-layer 2-head transformer variants. Each point $(x,y)$ represents the attention values $(\mathbb{A}^{H1}_{ij}, \mathbb{A}^{H2}_{ij})$ from heads 1 and 2 respectively for the same node pair $(i,j)$. Points lying on $x=y$ indicate identical attention patterns across heads, while deviations show head-specific specialization. We compare three architectures: SL (row 1, sparse learned attention), DLB (row 2, dense learned attention with graph bias), and DL (row 3, dense learned attention). Positive correlations suggest heads learn complementary rather than competing patterns.}
    \label{fig:Heads1L2H}
\end{figure}


\begin{figure}[h!]
    \centering
    \includegraphics[width=\textwidth]{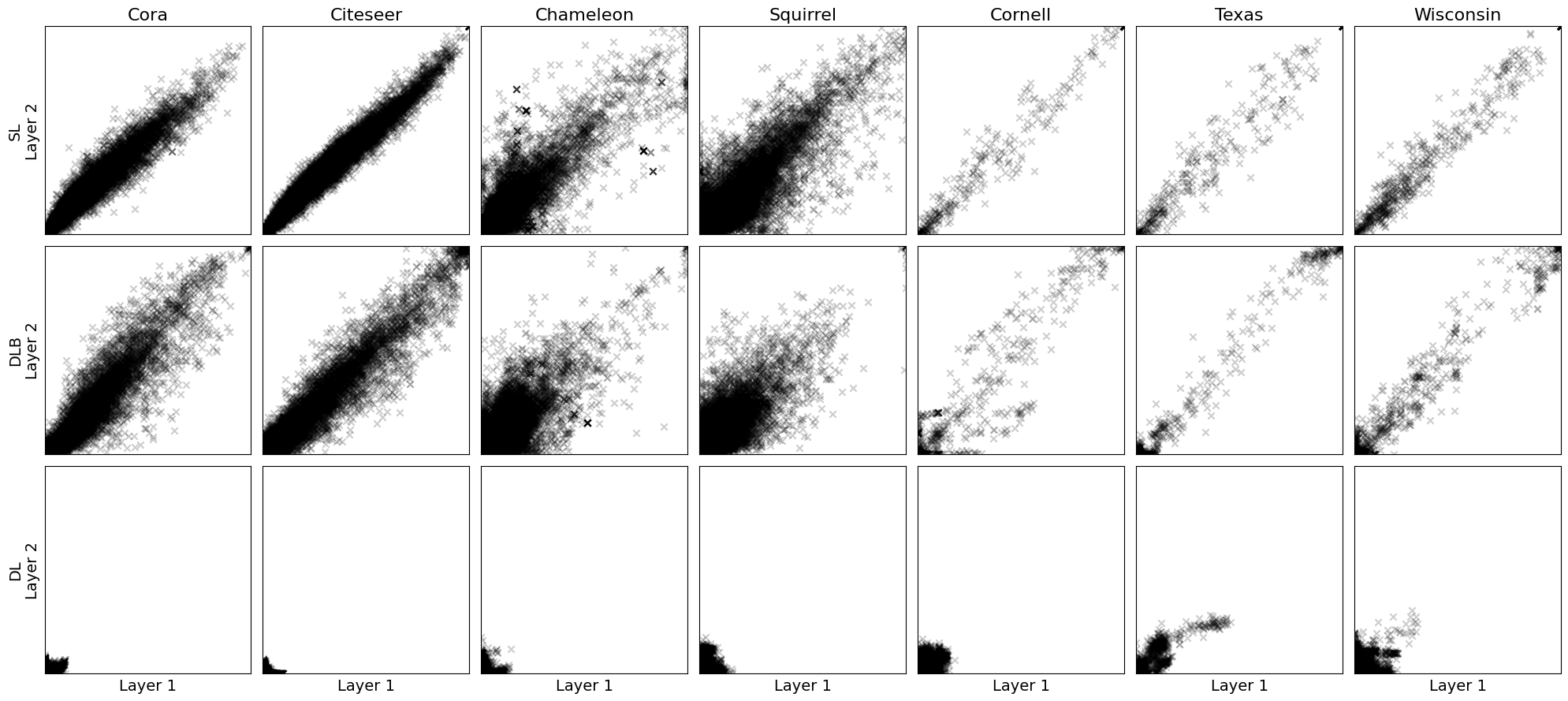}
    \caption{Comparison of attention patterns between layers in 2-layer 1-head transformer variants. Each point $(x,y)$ represents the attention values $(\mathbb{A}^{L1}_{ij}, \mathbb{A}^{L2}_{ij})$ from layers 1 and 2 respectively for the same node pair $(i,j)$. Points lying on $x=y$ indicate identical attention patterns across layers, while deviations show layer-specific specialization. We compare three architectures: SL (row 1, sparse learned attention), DLB (row 2, dense learned attention with graph bias), and DL (row 3, dense learned attention). Layer-wise attention patterns show weaker correlations compared to head-wise patterns (\Cref{fig:Heads1L2H}).}
    \label{fig:layers2L1H}
\end{figure}


\newpage
\section{Model Performance}
\begin{table}[h!]
\caption{Model accuracies across different architectures and datasets. For each architecture (SC: Sparse Constant, SL: Sparse Learned, DLB: Dense Learned with Bias, DL: Dense Learned), we report mean accuracies across 10 runs with standard deviations in parentheses. Results highlighted in \colorbox{lightblue}{blue} indicate best performing model(s) for each dataset. Training uses 40\% of nodes for training, 30\% for validation, and 30\% for testing. SC models only use 1 head since they have constant attention patterns.}
\label{tab:acc-model_comparison}
\centering
\begin{tabular}{llllllll}
\toprule
Model & Cora & Citeseer & Chameleon & Squirrel & Cornell & Texas & Wisconsin \\
\midrule
1L1H & & & & & & & \\
\hline
SC & \colorbox{lightblue}{\textbf{0.85}} & \colorbox{lightblue}{\textbf{0.75}} & 0.45 & 0.31 & 0.61 & 0.71 & 0.76 \\
    & (0.01) & (0.02) & (0.02) & (0.01) & (0.05) & (0.06) & (0.05) \\
SL & \colorbox{lightblue}{\textbf{0.85}} & 0.74 & \colorbox{lightblue}{\textbf{0.57}} & \colorbox{lightblue}{\textbf{0.40}} & 0.64 & \colorbox{lightblue}{\textbf{0.78}} & 0.79 \\
       & (0.01) & (0.02) & (0.02) & (0.01) & (0.03) & (0.06) & (0.05) \\
DLB & \colorbox{lightblue}{\textbf{0.85}} & \colorbox{lightblue}{\textbf{0.75}} & \colorbox{lightblue}{\textbf{0.57}} & \colorbox{lightblue}{\textbf{0.40}} & 0.65 & 0.77 & 0.80 \\
      & (0.01) & (0.01) & (0.02) & (0.02) & (0.05) & (0.07) & (0.05) \\
DL & 0.69 & 0.69 & 0.49 & 0.35 & \colorbox{lightblue}{\textbf{0.69}} & 0.75 & \colorbox{lightblue}{\textbf{0.81}} \\
         & (0.02) & (0.02) & (0.02) & (0.01) & (0.03) & (0.10) & (0.04) \\
\midrule
1L2H & & & & & & & \\
\hline
SL & \colorbox{lightblue}{\textbf{0.85}} & \colorbox{lightblue}{\textbf{0.74}} & \colorbox{lightblue}{\textbf{0.58}} & 0.40 & 0.66 & \colorbox{lightblue}{\textbf{0.78}} & 0.78 \\
 & (0.01) & (0.01) & (0.02) & (0.01) & (0.07) & (0.08) & (0.03) \\
DLB & 0.84 & \colorbox{lightblue}{\textbf{0.74}} & \colorbox{lightblue}{\textbf{0.58}} & \colorbox{lightblue}{\textbf{0.41}} & 0.67 & \colorbox{lightblue}{\textbf{0.78}} & 0.78 \\
 & (0.01) & (0.02) & (0.02) & (0.01) & (0.05) & (0.07) & (0.03) \\
DL & 0.69 & 0.69 & 0.50 & 0.36 & \colorbox{lightblue}{\textbf{0.71}} & \colorbox{lightblue}{\textbf{0.78}} & \colorbox{lightblue}{\textbf{0.80}} \\
 & (0.02) & (0.01) & (0.02) & (0.01) & (0.06) & (0.04) & (0.03) \\
\hline
2L1H & & & & & & & \\
\hline
SC & 0.86 & \colorbox{lightblue}{\textbf{0.75}} & 0.43 & 0.32 & 0.60 & 0.66 & 0.71 \\
 & (0.01) & (0.01) & (0.03) & (0.02) & (0.06) & (0.07) & (0.03) \\
SL & \colorbox{lightblue}{\textbf{0.87}} & 0.74 & 0.60 & \colorbox{lightblue}{\textbf{0.43}} & 0.64 & 0.77 & 0.77 \\
 & (0.01) & (0.01) & (0.02) & (0.01) & (0.05) & (0.07) & (0.03) \\
DLB & \colorbox{lightblue}{\textbf{0.87}} & \colorbox{lightblue}{\textbf{0.75}} & \colorbox{lightblue}{\textbf{0.61}} & \colorbox{lightblue}{\textbf{0.43}} & 0.63 & 0.73 & 0.76 \\
 & (0.01) & (0.01) & (0.01) & (0.02) & (0.09) & (0.08) & (0.05) \\
DL & 0.69 & 0.69 & 0.50 & 0.36 & \colorbox{lightblue}{\textbf{0.72}} & \colorbox{lightblue}{\textbf{0.79}} & \colorbox{lightblue}{\textbf{0.80}} \\
 & (0.02) & (0.01) & (0.02) & (0.01) & (0.06) & (0.07) & (0.03) \\
\hline
2L2H & & & & & & & \\
\hline
SL & \colorbox{lightblue}{\textbf{0.87}} & \colorbox{lightblue}{\textbf{0.74}} & 0.61 & \colorbox{lightblue}{\textbf{0.44}} & 0.61 & \colorbox{lightblue}{\textbf{0.77}} & 0.75 \\
 & (0.01) & (0.01) & (0.02) & (0.01) & (0.07) & (0.07) & (0.02) \\
DLB & 0.86 & \colorbox{lightblue}{\textbf{0.74}} & \colorbox{lightblue}{\textbf{0.63}} & \colorbox{lightblue}{\textbf{0.44}} & 0.62 & 0.75 & 0.78 \\
 & (0.01) & (0.01) & (0.02) & (0.01) & (0.09) & (0.07) & (0.04) \\
DL & 0.69 & 0.69 & 0.50 & 0.36 & \colorbox{lightblue}{\textbf{0.70}} & \colorbox{lightblue}{\textbf{0.77}} & \colorbox{lightblue}{\textbf{0.81}} \\
 & (0.02) & (0.01) & (0.02) & (0.01) & (0.06) & (0.08) & (0.03) \\
\hline
\end{tabular}
\end{table}


\newpage
\section{Attention Patterns}
\begin{table}[h!]
    \caption{Comparing the relative attention distribution between local and global neighborhoods. For each model architecture and dataset, we compute the ratio of average attention paid to non-neighbors versus neighbors. A ratio of 0 indicates attention is strictly local (only within defined graph neighborhoods), while 1 indicates uniform attention distribution across all nodes regardless of connectivity. Ratios > 1 suggest the model preferentially attends to nodes outside the local neighborhood. Self-attention is considered as within-neighborhood attention.}
        \label{tab:attention-ratio}
\centering
\resizebox{0.9\linewidth}{!}{
\begin{tabular}{llccccccc}
\toprule
& \textbf{Model} & \textbf{Cora} & \textbf{Citeseer} & \textbf{Chameleon} & \textbf{Squirrel} & \textbf{Cornell} & \textbf{Texas} & \textbf{Wisconsin} \\
\midrule
\multirow{3}{*}{\textbf{1L1H}} & SL & 0.0  & 0.0      & 0.0       & 0.0      & 0.0     & 0.0   & 0.0\\
& DLB & 0.0  & 0.0      & 0.01      & 0.0      & 0.01    & 0.0   & 0.0     \\
& DT  & 1.13 & 0.68     & 1.03      & 1.16     & 1.28    & 1.21  & 0.86      \\ 
\midrule
\multirow{3}{*}{\textbf{1L2H}} &SL & 0.00 & 0.00 & 0.00 & 0.00 & 0.00 & 0.00 & 0.00  \\
& DLB & 0.00 & 0.00 & 0.00 & 0.00 & 0.00 & 0.00 & 0.00  \\
& DL & 0.79 & 0.55 & 0.97 & 1.14 & 0.86 & 1.16 & 0.91  \\
\midrule
\multirow{3}{*}{\textbf{2L1H}} &SL & 0.00 & 0.00 & 0.00 & 0.01 & 0.00 & 0.00 & 0.00 \\
& DLB & 0.00 & 0.00 & 0.02 & 0.02 & 0.01 & 0.01 & 0.01  \\
& DL & 0.96 & 0.77 & 0.94 & 1.25 & 1.00 & 0.75 & 0.98  \\
\midrule
\multirow{3}{*}{\textbf{2L2H}} &SL & 0.00 & 0.00 & 0.00 & 0.01 & 0.00 & 0.00 & 0.00 \\
& DLB & 0.00 & 0.00 & 0.02 & 0.03 & 0.01 & 0.01 & 0.01  \\
& DL & 0.89 & 0.84 & 1.14 & 1.12 & 0.91 & 0.77 & 0.88 \\
\bottomrule
\end{tabular}
}
\end{table}


\begin{figure}[h!]
    \centering
        \includegraphics[width=0.95\linewidth]{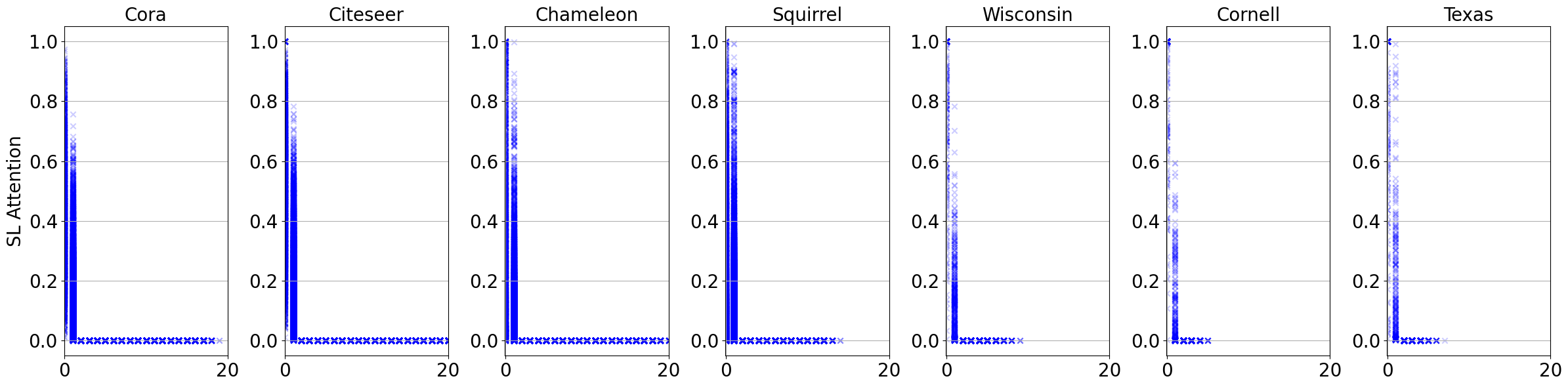}
        \includegraphics[width=0.95\linewidth]{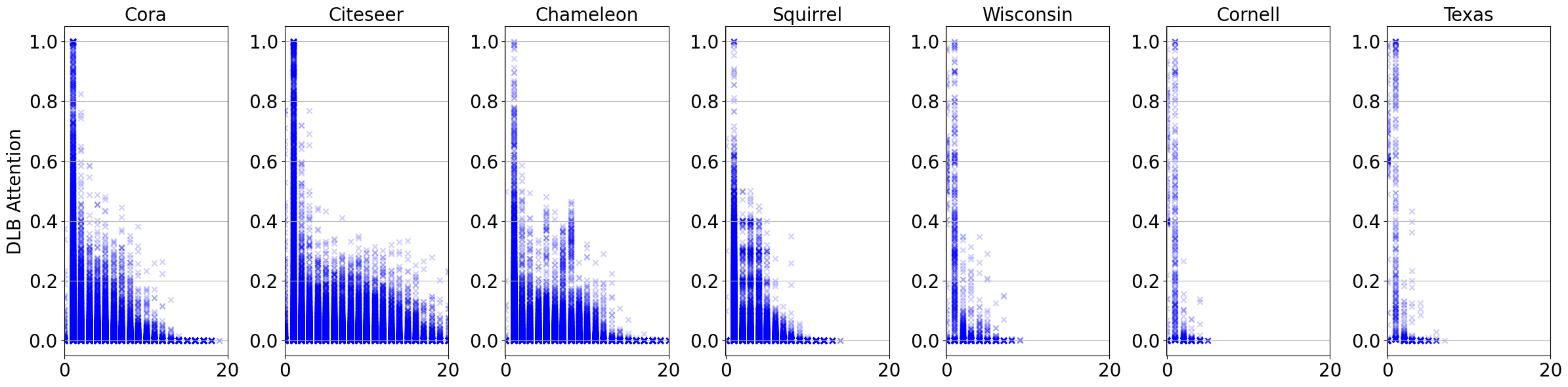}
        \includegraphics[width=0.95\linewidth]{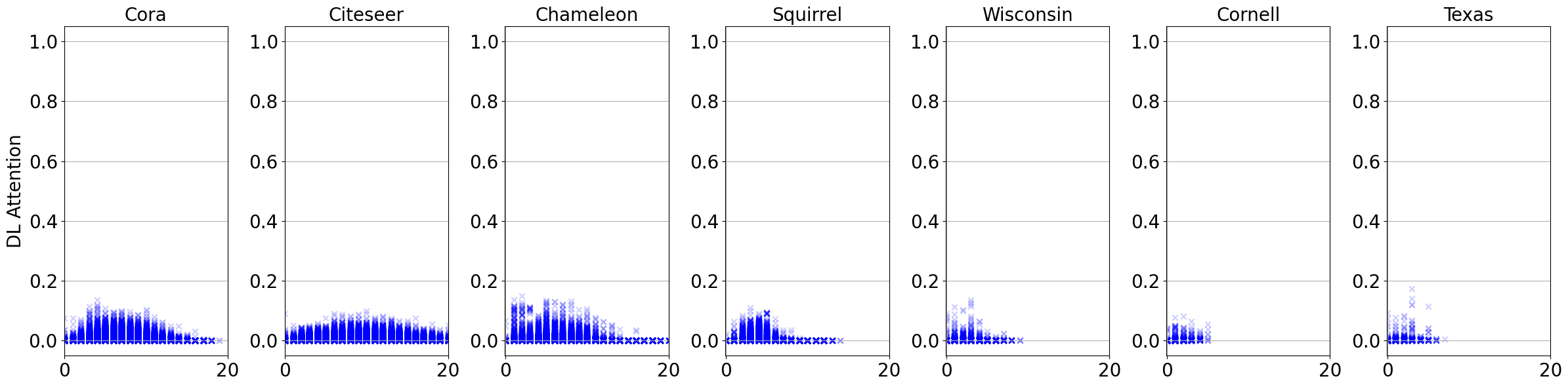}
        \caption{Distribution of attention values across n-hop neighborhoods in 1-layer 1-head models for SL (row 1), DLB (row 2), and DL (row 3). Each blue marker \textcolor{blue}{$\times$} represents how much attention a node pays (y-axis) to nodes at different hop distances (x-axis). For example, if node $i$ pays 0.5 attention to node $j$ which is 3 hops away, this is plotted as point $(3, 0.5)$. Self-attention (node attending to itself) is considered as 0-hop. Note how DLB models (row 2) focus attention on closer nodes while DL models (row 3) distribute attention more uniformly across all hop distances.}
\label{fig:attention-nhop-distribution}
\end{figure}



\section{Extended \Cref{fig:mechanisms}}

\begin{figure}[h!] 
    \centering
    \includegraphics[width=0.97\textwidth]{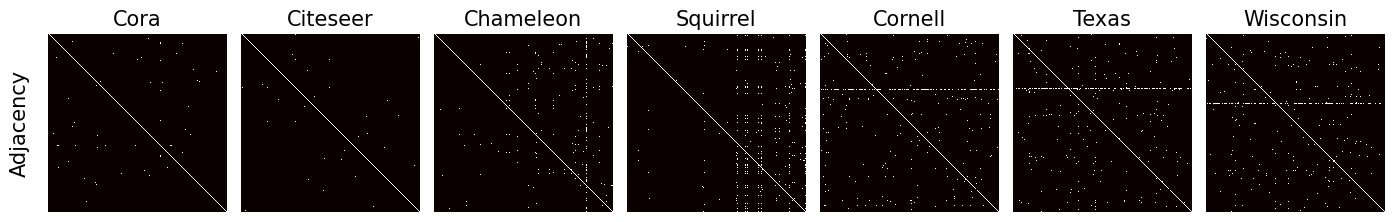}
    \includegraphics[width=\textwidth]{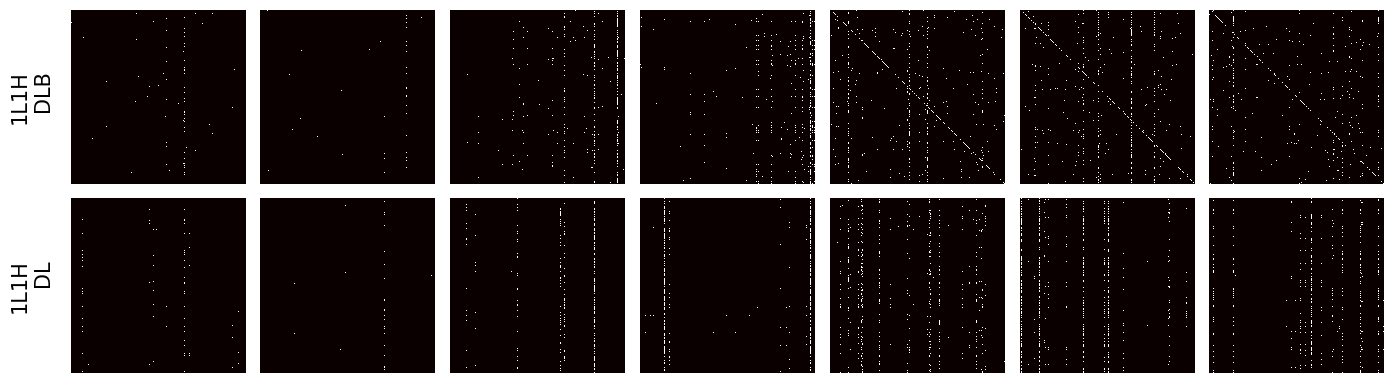}
    \includegraphics[width=\textwidth]{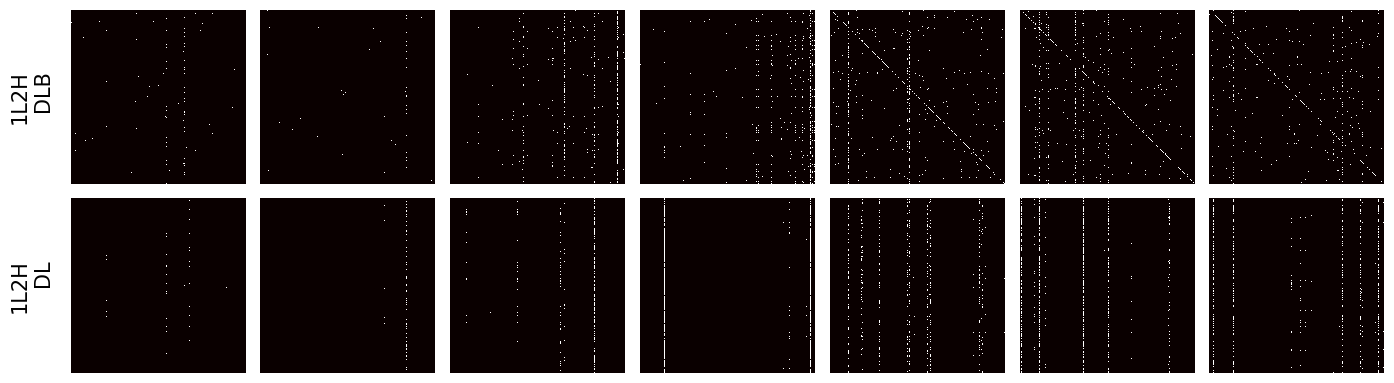}
    \includegraphics[width=\textwidth]{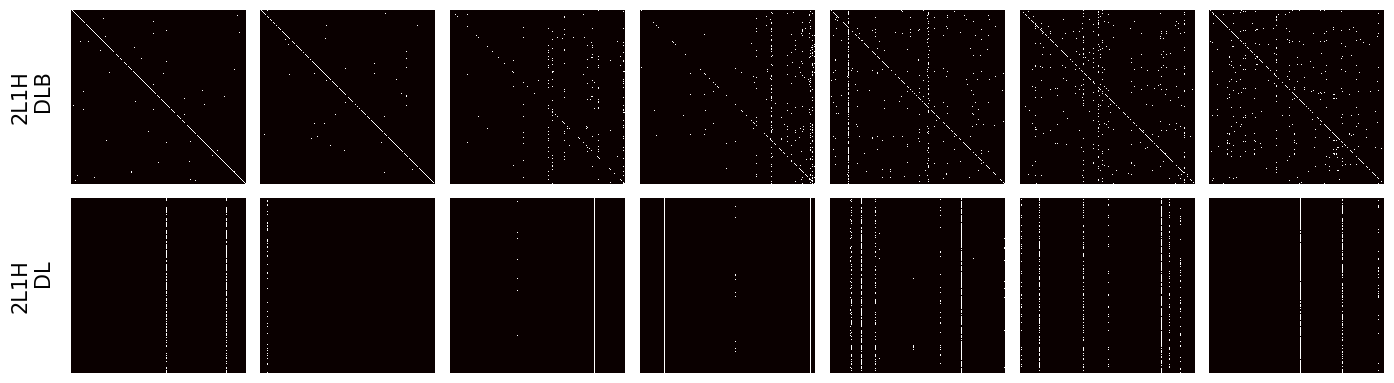}
    \includegraphics[width=\textwidth]{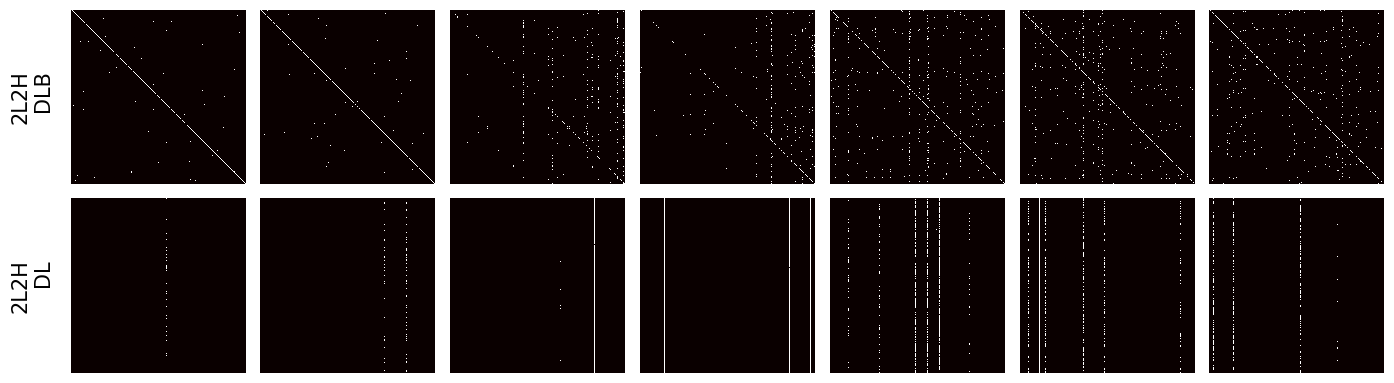}
    \caption{\textbf{Comparing adjacency and quasi-adjacency matrices across architectures and configurations.} 
    Row 1: Original adjacency matrices for each dataset. 
    Rows 2-9: Quasi-adjacency matrices constructed by thresholding Attention Graphs (see \Cref{apdx:recovering-graph-structure}) for different model configurations. 
    These visualizations reveal distinct information flow patterns and algorithmic strategies learned by different architectures despite similar performance.
    DLB models exhibit strong self-attention patterns (diagonal lines), suggesting they focus on initial node features rather than aggregating information from neighbors.
    DL models develop reference nodes (vertical lines) that receive high attention from all other nodes, suggesting a classification algorithm based on comparing nodes against these references.
    Extended version of \Cref{fig:mechanisms}.
    }
    \label{fig:HM}
\end{figure}


\end{document}